\documentclass[11pt]{article}
\usepackage{amsmath}
\usepackage{authblk}
\usepackage{bm}
\usepackage{booktabs}
\usepackage{caption}
\usepackage{enumitem}
\usepackage[margin=1in]{geometry}
\usepackage{graphicx}
\usepackage{mathrsfs}
\usepackage{multirow}
\usepackage{subcaption}
\usepackage[colorlinks=true,citecolor=blue]{hyperref}
\usepackage[numbers,sort&compress]{natbib}

\DeclareUnicodeCharacter{2009}{ }

\title{Physics-informed deep learning for infectious disease forecasting}
\author[1]{Ying Qian}
\author[1]{Kui Zhang}
\author[2]{\'{E}ric Marty}
\author[2]{Avranil Basu}
\author[2]{Eamon B. O’Dea}
\author[3]{Xianqiao Wang}
\author[4]{Spencer Fox}
\author[2,5,6]{Pejman Rohani}
\author[2*]{John M. Drake}
\author[1*]{He Li}
\affil[1]{School of Chemical, Materials, and Biomedical Engineering, University of Georgia, Athens, GA 30602}
\affil[2]{Odum School of Ecology and Center for the Ecology of Infectious Diseases, University of Georgia, Athens,
GA 30602}
\affil[3]{School of Environmental, Civil Agricultural and Mechanical Engineering, College of Engineering, University of Georgia, Athens, GA, 30602}
\affil[4]{Department of Epidemiology \& Biostatistics, University of Georgia, Athens, GA, 30602}
\affil[5]{Department of Infectious Diseases, College of Veterinary Medicine, University of Georgia, Athens,
GA 30602}
\affil[6]{Center for Influenza Disease and Emergence Research, University of Georgia, Athens,
GA 30602}
\affil[*]{jdrake@uga.edu, he.li3@uga.edu}
\date{}  

\begin{document}

\maketitle

\begin{abstract}

Accurate forecasting of contagious illnesses has become increasingly important to public health policymaking, and better prediction could prevent the loss of millions of lives.  To better prepare for future pandemics, it is essential to improve forecasting methods and capabilities. In this work, we propose a new infectious disease forecasting model based on physics-informed neural networks (PINNs), an emerging area of scientific machine learning. The proposed PINN model incorporates dynamical systems representations of disease transmission into the loss function, thereby assimilating epidemiological theory and data using neural networks (NNs). Our approach is designed to prevent model overfitting, which often occurs when training deep learning models with observation data alone. In addition, we employ an additional sub-network to account for mobility, cumulative vaccine doses, and other covariates that influence the transmission rate, a key parameter in the compartmental model. To demonstrate the capability of the proposed model, we examine the performance of the model using state-level COVID-19 data in California. Our simulation results show that predictions of PINN model on the number of cases, deaths, and hospitalizations are consistent with existing benchmarks. In particular, the PINN model outperforms naive baseline forecasts and various sequence deep learning models, such as Recurrent Neural Networks (RNNs), Long Short-Term Memory (LSTM) networks, Gated Recurrent Units (GRUs), and Transformer models. We also show that the performance of the PINN model is comparable to a sophisticated Gaussian infection state forecasting model that combines the compartmental model, a data observation model and a regression model for inferring parameters in the compartmental model. Nonetheless, the PINN model offers a simpler structure and is easier to implement. In summary, we perform a systematic study of the predictive capability of the PINN model in forecasting the dynamics of infectious diseases and our results showcase the potential of the proposed model as an efficient computational tool to enhance the current capacity of infectious disease forecasting.


\end{abstract}

\section{Introduction}

COVID-19 was the third largest cause of mortality in the United States in 2022~\cite{ahmad2022provisional} and, in addition to Ebola, Chikungunya, Zika, and mPox, was one of five global outbreaks in the past decade caused by emerging pathogens. Climate change is expected to exacerbate disease transmission risks, which before COVID-19 already accounted for roughly 25\% of global mortality~\cite{morens2004challenge,carlson2022climate}. In the midst of an exponentially growing outbreak, public health officials must respond rapidly in the face of intense uncertainty~\cite{biggerstaff2022improving}. Recent developments in infectious disease forecasting have sought to improve our ability to anticipate future epidemiological trends, such as the number of reported cases, deaths, or hospitalizations from a disease, and are increasingly used during outbreaks. Forecasting provides a concrete way to integrate data with epidemiological knowledge and can be used to inform the allocation of critical resources such as antiviral drugs or ventilators, the implementation of non-pharmaceutical interventions, and the design of vaccine trials~\cite{biggerstaff2022improving,lutz2019applying,case2025accurate}.

Compartmental epidemiological models, which divide the population into distinct mutually exclusive subsets based on individuals' disease status, have been widely employed to represent and forecast the spread of infectious diseases within a population~\cite{Anderson&May:1991,Keeling&Rohani:2008,brauer2008compartmental,brauer2019mathematical}.  The most basic compartments include individuals who are susceptible to the infection (Susceptible ($S$)), those who are currently infected and capable of transmitting the disease (Infected ($I$)), and individuals who have recovered from the infection and are assumed to have acquired immunity or those who have been removed from the susceptible pool due to death (Recovered or Removed ($R$))~\cite{martcheva2015introduction,huppert2013mathematical}. Numerous extensions of the basic compartmental models have introduced more compartments to account for other state variables, such as death or variations in disease staging~\cite{Wearing:2005p596,massonis2021structural,biala2021fractional,ndairou2020mathematical,abou2020compartmental,batistela2021sirsi,odagaki2023new} or other flows among compartments (\textit{e.g.}, loss of immunity~\cite{Keeling&Rohani:2008,Domenech_et:2018}). The transitions between these compartments are governed by a set of ordinary differential equations that describe the rates at which individuals move from one compartment to another. Although compartmental epidemiological models have provided valuable insights into the general dynamics of infectious diseases and inform public health strategies, compartmental models typically assume parameters such as the transmission and recovery rates are fixed.  As has been observed in recent high-profile outbreaks, this assumption does not adequately capture the dynamic nature of an unfolding epidemic, especially when factors like the rollout of new vaccines, public health interventions, or virus mutation are involved (see~\cite{Kucharski:2015ff,o2022semi, drake2023data,gibson2020real}).

\begin{figure}[ht]
  \centering
\hspace{0.1in}\includegraphics[width=0.95\textwidth]{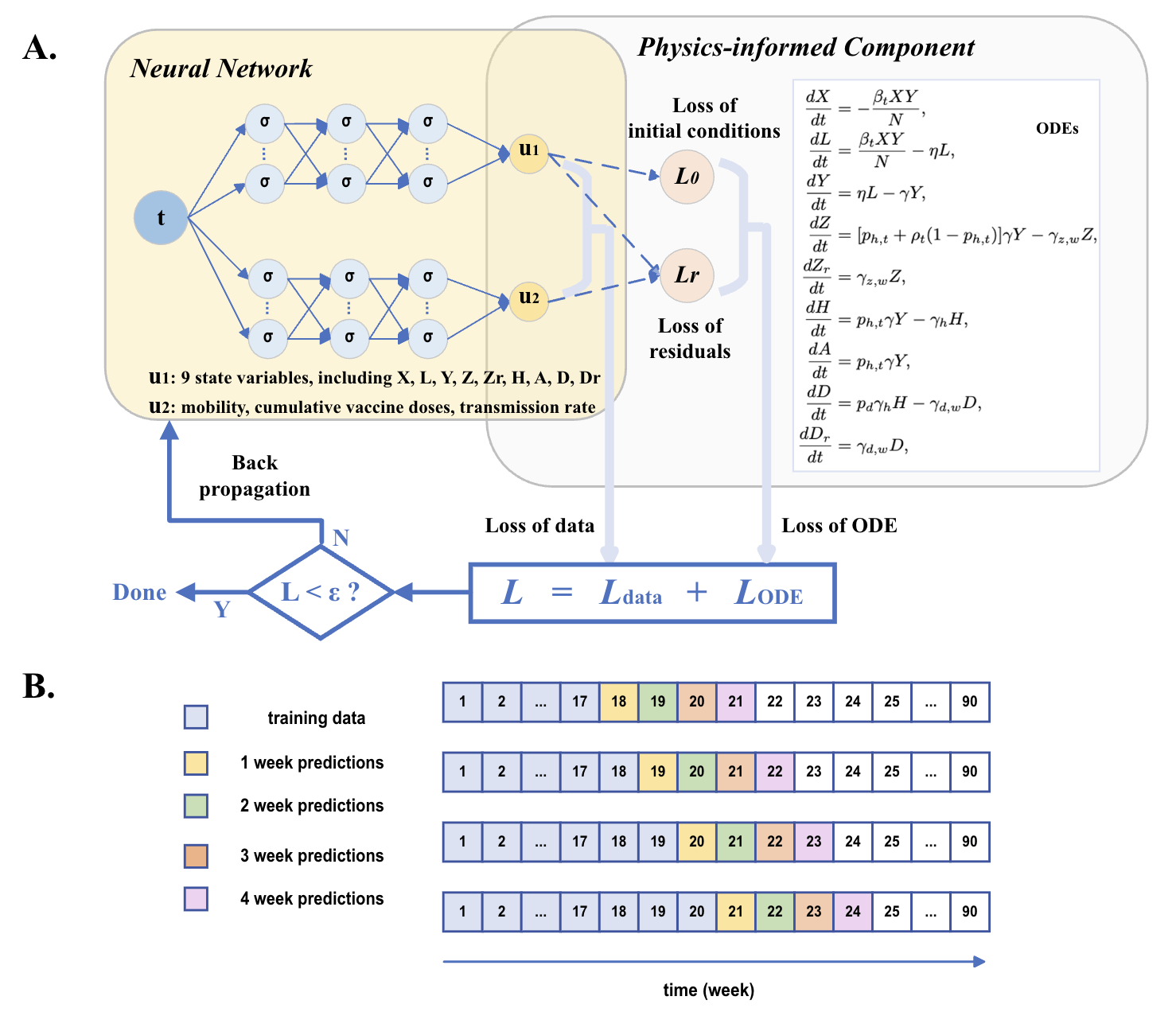}
  \caption{\textbf{A}. Schematic of the proposed PINNs model for infectious disease forecasting. The model comprises two sub-networks: the upper sub-network predicts the state variables in the compartmental model, while the lower sub-network estimates the time-dependent model parameters. The output $u_1$ represents the nine compartmental state variables, including $X$, $L$, $Z$, $Z_r$, $H$, $A$, $D$, $D_r$. The output $u_2$ represents factors including mobility, cumulative vaccine doses, and transmission rate. The data loss $L_{data}$ is a weighted sum of observable state variables and factors. \textbf{B}. Schematic of the rolling window approach. As training data accumulates over time, the PINNs model continuously updates and generates forecasts for the subsequent 1–4 weeks.}
  \label{figure_pinn_schematic}
\end{figure}

Over the past decade, alternative approaches to tackling complex social problems have emerged, including artificial intelligence (AI) and machine learning (ML).   These approaches have the advantage of efficiently identifying patterns in data and can, in principle, accommodate multiple disparate data streams. However, conventional AI/ML methods require a large amount of training and labeled data to ensure good performance~\cite{wu2018deep,deng2023deep,lu2023systematic,gao2023deep}. Furthermore, AI/ML models may provide unrealistic predictions, especially for non-stationary systems, as they are not constrained by the mechanisms driving transmission. Furthermore, data-driven models have historically struggled with predicting epidemics because they often fail to differentiate real trends from noise in the data collecting process~\cite{vytla2021mathematical}. Recently, AI-based solvers for partial differential equations (PDEs) and ordinary differential equations (ODEs) have attracted growing attention due to their ability to assimilate the underlying scientific laws represented by ODEs or PDEs with the learning properties of neural networks (NNs). Combining AI with dynamical systems models can help to infer unknown model parameters as constants or functions using limited data. This enables learning from ``small data'' as we explicitly utilize the constraints from the physical or biological laws~\cite{karniadakis_2021,raissi2020hidden,lu2020extraction,zou2025multi,zou2023correcting,zhang2023discovering,zou2023uncertainty,zou2022neuraluq,yang2021b,meng2020ppinn,yu2022gradient,lou2021physics,lu2021deepxde,toscano2025pinns}. Physics-informed neural networks (PINNs)~\cite{raissi2019physics} provide a powerful framework that integrates data with compartmental models by incorporating both into the loss function. As illustrated in Fig.\ref{figure_pinn_schematic}A, the left side of the architecture represents a fully connected neural network (FCNN), corresponding to the physics-uninformed component, while the right side captures the physics-informed component. The loss function is designed to account for contributions from both observed data and the underlying compartmental models. PINNs have been successfully applied to a range of scientific and engineering challenges, encompassing both forward and inverse problems~\cite{zhang2022aoslo,chen2023tgm,daneker2024transfer,chen2024deep,qian2024coagulo,cai2021artificial,linka2022bayesian,patel2022thermodynamically,patel2021physics, zou2025learning}.

Here, we introduce a novel disease forecasting model based on PINNs. The compartmental model includes nine state variables (Tab.~\ref{table_var_stat}), with the PINNs architecture comprising two sub-networks that take time \( t \) as input. The upper sub-network predicts all nine state variables, contributing to the ODE loss \( \mathcal{L}_{\text{ODE}} \), while three observable variables -- reported cases (\( Z_r \)), hospitalizations (\( A \)), and reported deaths (\( D_r \)) -- also contribute to the data loss \( \mathcal{L}_{\text{data}} \). The lower sub-network captures the relationship between transmission rate and two associated factors: mobility and cumulative vaccine doses. It outputs predictions for all three, with transmission rate feeding into \( \mathcal{L}_{\text{ODE}} \), and the other two informing \( \mathcal{L}_{\text{data}} \). This design allows for seamless integration of mechanistic modeling with observed data, enhancing forecasting performance.

To demonstrate the capability of this approach, we evaluated the performance of the model by comparing 1-, 2-, 3- and 4-weeks-ahead forecasts of COVID-19 cases, deaths, and hospitalizations in the state of California using reports of COVID-19, Google’s mobility reports, and vaccination data available each week. We assessed model accuracy on both point forecasting and quantile forecasting~\cite{bracher2021evaluating}. We compared the score of our model with a naive baseline forecast, various sequence deep learning models, such as recurrent neural networks (RNNs), long short-term memory
(LSTM) networks, gated recurrent units (GRUs), and Transformer models as well as a traditional mathematical model.

\section{Data and Methods}

\subsection{Data}

\subsubsection{Input Data}

\begin{figure}[ht]
  \centering
\hspace{0.1in}\includegraphics[width=\textwidth]{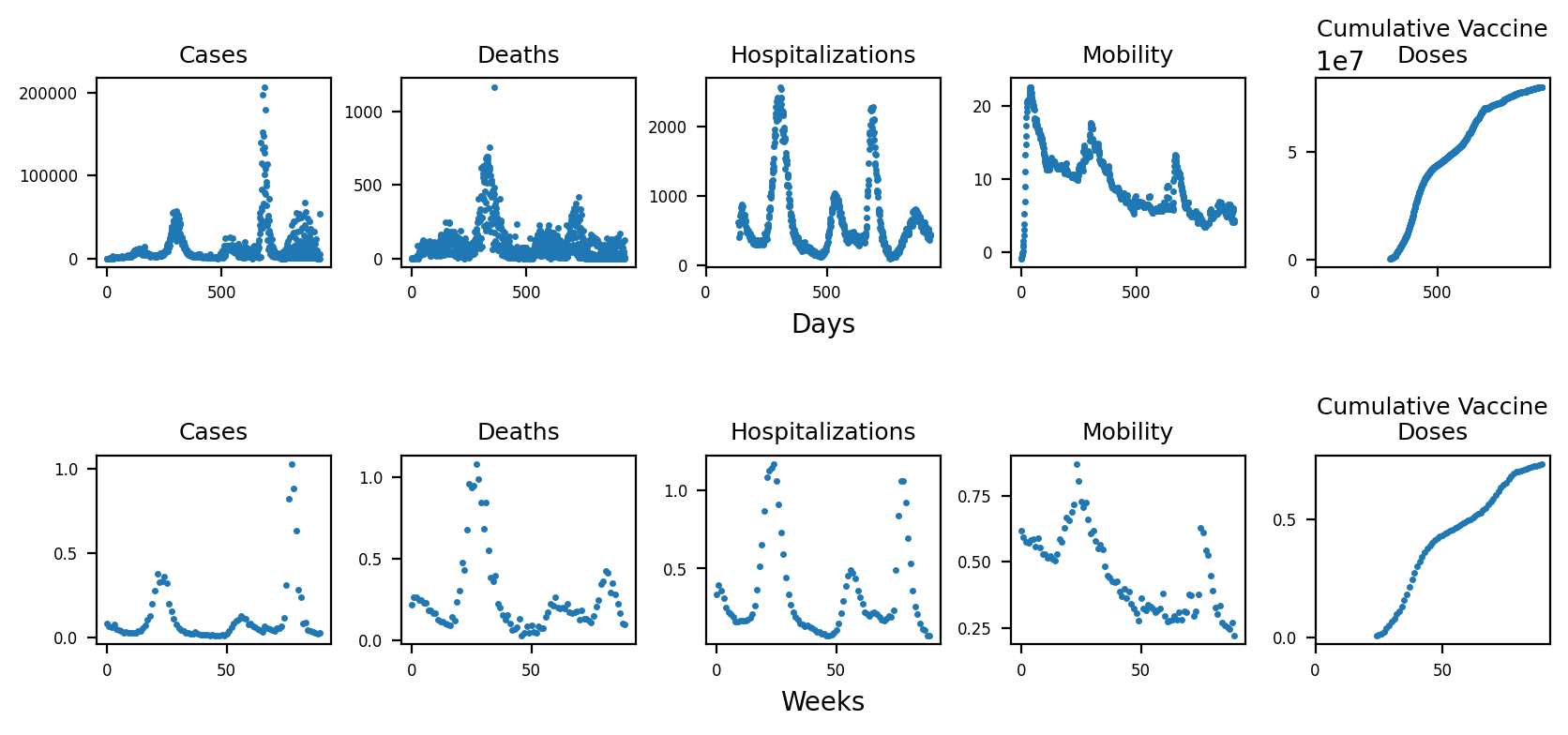}
  \caption{Original dataset (upper panel) and preprocessed (lower panel) dataset used for training and testing the proposed PINNs model.} 
  \label{figure_data}
\end{figure}

To facilitate a comparison of our model's forecasting performance with that of various other models, we used the forecasting targets defined by the COVID-19 Forecast Hub~\cite{ray2020ensemble, o2022semi}, including COVID-19 case and death data sourced from the COVID-19 Data Repository available at \url{https://github.com/CSSEGISandData/COVID-19}, which is maintained by the Center for Systems Science and Engineering (CSSE) at Johns Hopkins University, Baltimore, MD, USA (JHU)~\cite{dong2020interactive}.  Furthermore, we used information on hospitalizations from the \textit{COVID-19 Reported Patient Impact and Hospital Capacity by State Timeseries} and \textit{COVID-19 Reported Patient Impact and Hospital Capacity by State} datasets compiled by the US Department of Health \& Human Services (HHS) available on healthdata.gov. Access to these datasets was facilitated through the Delphi Epidata API, as detailed by Reinhart et al.~\cite{reinhart2021open}. For our analysis, the time series of hospitalizations is the sum of the fields labeled previous\_day\_admission\_adult\_covid\_confirmed and previous\_day\_admission\_pediatric\_covid\_confirmed in the tables provided by HHS. 

We used data on mobility and cumulative vaccine doses due to their association with the transmission rate, a critical time-dependent parameter in epidemiological forecasting. In particular, mobility can be measured by the amount of time individuals spend in residential areas, as quantified in Google’s community mobility reports~\cite{google2020google}. For a given forecast date, we obtained the latest snapshot of Google’s \textit{Global\_Mobility\_Report.csv} file on \url{http://web.archive.org} that was made before the beginning forecast date in the UTC time zone.

Cumulative vaccine doses represent the total number of vaccine doses administered within a given state. Our modeling assumes that the average susceptibility of individuals who have not been previously infected decreases as the number of administered doses increases. Rather than modeling the number of susceptible individuals directly, we model the effect of vaccination on average susceptibility, acknowledging that a single dose of a COVID-19 vaccine typically does not confer full immunity. We obtained the time series data on administered vaccine doses from the GitHub repository maintained by Johns Hopkins University (\url{www.github.com/govex/COVID-19}). This dataset compiles information from state public dashboards and the Centers for Disease Control and Prevention (CDC) Vaccine Tracker. For each day, the dataset reports the higher value between the two sources to provide the most current estimate. To generate a covariate without missing values, we assumed all values prior to the first reported entry were zero, and any missing values within the series were imputed using linear interpolation.

\subsubsection{Data preprocessing}

The time series data for cases, deaths, hospitalizations, mobility, and cumulative vaccine doses were originally provided at a daily resolution, whereas our prediction targets are weekly totals for cases, deaths, and hospitalizations. To ensure consistency with the prediction targets, we preprocessed these datasets by computing a 7-day moving average and aggregating the daily values into a weekly time series. This smoothing strategy not only aligns the temporal resolution with the forecasting targets but also reduces short-term fluctuations in the original data.

The hospitalization data differ from the other time series in several key aspects. First, the hospitalization records were made available later than the case and death data, with reliable reporting beginning in July 2020. Second, the initial portion of the hospitalization time series is affected by an artificial upward trend caused by the gradual increase in the number of hospitals reporting data. To mitigate this bias, we excluded the first 20 weeks of hospitalization data from our analysis.

Toward the end of the time series, a small, incomplete peak appears following two major peaks. To maintain focus on well-characterized dynamics, we chose to model only the two major peaks and discarded data after week 110. These two peaks correspond to different viral variants (Beta and Delta), each associated with distinct transmission dynamics, and are sufficient for evaluating the performance of our forecasting model.

Fig.~\ref{figure_data} illustrates both the original (upper panel) and preprocessed (lower panel) time series data. To address the large differences in scale across variables, we applied linear scaling to normalize each time series to a similar range, approximately between 0 and 1. This normalization is also reflected in the lower panel of Fig.~\ref{figure_data}, facilitating better visualization and more stable model training.

\subsection{Physics-informed neural networks}

\begin{table}[ht]
    \centering
    \begin{tabular}{| c | p{12cm} |}
    \hline
    Name & Definition \\
    \hline
    $X$ & Uninfected and susceptible individuals \\ 
    $L$ & Individuals with latent infections who are not yet infectious \\  
    $Y$ & Individuals who are infectious \\  
    $Z$ & Individuals who have been diagnosed and will be reported as cases but have not yet been reported \\  
    $Z_r$ & This compartment keeps track of the number of cases reported each day \\  
    $H$ & Individuals who have been hospitalized \\ 
    $A$ & Individuals who are new hospital admissions \\  
    $D$ & Individuals who have died from the infection but whose death has not yet been reported \\  
    $D_r$ & The number of newly reported deaths each week \\ 
    \hline
    \end{tabular}
    \caption{State variables in the employed compartmental model by O’Dea and Drake~\cite{o2022semi}.}
    \label{table_var_stat}
\end{table}

Physics-informed neural networks (PINNs), which combine traditional compartmental models with observational data through neural networks, have emerged as one of the most influential approaches in scientific machine learning. In this work, we incorporate the epidemiological model proposed in~\cite{o2022semi} -- a compartmental model for COVID-19 with time-dependent parameters -- into our PINNs framework. This model is governed by a system of nine ODEs, involving nine state variables and eleven parameters. The definitions of the state variables are provided in Tab.~\ref{table_var_stat}. Among these, three state variables -- reported cases $Z_r$, reported deaths $D_r$, and hospitalizations $A$ -- are observable and used as targets for model training and evaluation. The remaining six state variables are unobservable and are treated as latent variables within the PINNs framework.

\begin{table}[ht]
    \centering
    \setlength{\tabcolsep}{10pt} 
    \renewcommand{\arraystretch}{1.2} 
    \begin{tabular}{ | c | p{10cm} | c| }
    \hline
    Name & Definition & Value\\
    \hline
    $\beta_t$ & Transmission rate & estimated \\ 
    $N$ & Population size & 39512223 \\ 
    $\eta$ & Incubation rate & 0.25 \\ 
    $\gamma$ & Removal rate & 0.25 \\ 
    $\gamma_{d,w}$ & Rate of reporting deaths on day of week w & 0.1 \\ 
    $\gamma_{z,w}$ & Rate of reporting cases on day of week w & 1 \\
    $\gamma_h$ & Rate of exit from hospital & 0.1 \\ 
    $p_{h,t}$ & Probability of an infection leading to hospitalization & estimated \\ 
    $p_d$ & Probability of a hospitalization leading to death & estimated \\ 
    $\rho_t$ & Probability of a removal on day $t$ being reported as a case & 0.5 \\
    \hline
    \end{tabular}
    \caption{Model parameters used in the compartmental model by O’Dea and Drake~\cite{o2022semi}.}
    \label{table_param}
\end{table}

The system of ODEs is incorporated into the physics-informed component of the model, as illustrated in Fig.~\ref{figure_pinn_schematic}A. Definitions and values for each model parameter are provided in Tab.~\ref{table_param}. For consistency and comparability, we generally adopt the parameter settings from~\cite{o2022semi}: seven parameters are treated as known and constant, while three are estimated during training. Among them, the transmission rate $\beta_t$ is considered a critical time-dependent parameter and is predicted by the lower sub-network, as depicted in Fig.~\ref{figure_pinn_schematic}A. The remaining two parameters, $p_h$ and $p_d$, are assumed to be constant and are integrated into the model parameters to be estimated during the training process.

As depicted in Fig.~\ref{figure_pinn_schematic}A, the PINNs architecture comprises two sub-networks, both taking time $t$ as input. Each sub-network consists of three hidden layers with 50 neurons in each layer. The upper sub-network predicts the values of the nine state variables, all of which are used to compute the ODE loss, denoted as $\mathcal{L}_{ODE}$. Among these, three variables -- reported cases $Z_r$, hospitalizations $A$, and reported deaths $D_r$ -- are observable and contribute to the data loss $\mathcal{L}_{data}$.

The lower sub-network is designed to implicitly capture the relationship between the transmission rate and two closely associated factors: mobility and cumulative vaccine doses. It outputs predictions for mobility, cumulative vaccine doses, and the transmission rate. The transmission rate is a key time-dependent parameter in the compartmental model and is involved in the computation of $\mathcal{L}_{ODE}$, while mobility and cumulative vaccine doses has corresponding reported data and contribute to $\mathcal{L}_{data}$. This PINNs framework enables the integration of mechanistic modeling with real-world data for improved forecasting accuracy.

The constraint imposed by the compartmental model was integrated into the PINNs training process by combining the ODE loss with the data loss. Specifically, the total loss is defined as a weighted sum of the data loss and the ODE loss:

\begin{align}
    \mathcal{L}(\theta) = \mathcal{L}_{data}(\theta) + w_{ODE} \mathcal{L}_{ODE}(\theta)
\end{align}
where $\theta$ denotes the PINNs parameters, $\mathcal{L}_{data}$ represents data loss, $\mathcal{L}_{ODE}$ represents ODE loss, and $w_{ODE}$ represents the weight for ODE loss.

We denote the PINNs model as \( \bm{y}_\theta(\bm{t}) \), where \( \bm{t} \) represents the input time, and \( \bm{y} = [y_1, y_2, y_3, y_4, y_5]^T \) denotes the output vector, with \( y_1 \) to \( y_5 \) corresponding to cases, deaths, hospitalizations, mobility, and cumulative vaccine doses, respectively. We used the mean squared error (MSE) as the training metric, with the following loss function for the data:

\begin{equation}
    \mathcal{L}_{data}(\theta) = \frac{1}{N_{data}}\sum_{i=1}^{N_{data}} |\bm{y}_\theta(\bm{t}_i) - \bm{y}_i|^2,
\end{equation}
where \( N_{\text{data}} \) denotes the number of training data points, and \( \{\bm{t}_i, \bm{y}_i\} \) represents the \( i \)th data pair. The PINNs \( \bm{y}_\theta \) were trained by minimizing the loss function \( \mathcal{L}(\theta) \).

The ODE loss was defined as the sum of the loss on the initial conditions and the loss on the residuals.
\begin{equation}
    \mathcal{L}_{ODE}(\theta) = \mathcal{L}_0(\theta) + \mathcal{L}_r(\theta)
\end{equation}
where $\mathcal{L}_{0}$ represents initial condition loss, $\mathcal{L}_{r}$ represents residual loss.

Initial condition loss $\mathcal{L}_0(\theta)$ and residual loss $\mathcal{L}_r(\theta)$ were measured by MSE:
\begin{gather}
    \mathcal{L}_0(\theta) = \frac{1}{N_0} \sum_{i=1}^{N_0}{|\bm{y}_\theta(\bm{t}_0)-\bm{y_i}|^2} \\ 
    \mathcal{L}_r(\theta) = \frac{1}{N_{ODE}} \sum_{i=1}^{N_{ODE}}{|w_if(t_i)|^2}
\end{gather}
where $N_0$ denotes the number of initial data, $N_{ODE}$ denotes the number of ODEs, $w_i$ denotes the weight for each ODE, and $f(t)$ is defined to be given by the difference between left-hand-side and right-hand-side of ODE; \textit{i.e.}, for the first equation $\frac{dX}{dt} = (-\frac{\beta_t X Y}{N})$ in the ODE system, 
\begin{equation}
    f := \frac{dX}{dt} - (-\frac{\beta_t X Y}{N})
\end{equation}

Overfitting is a common challenge in time series forecasting. To mitigate this, we employ L2 regularization, a widely used technique that adds a penalty term -- the sum of the squared network parameters -- to the original loss function during training. This regularization discourages overly complex models by penalizing large parameter values. In practice, L2 regularization is well-integrated in \texttt{PyTorch} and can be easily applied by setting the \texttt{weight\_decay} parameter in the optimizer.

We used the Adam optimizer~\cite{kingma2014adam} with a learning rate of \( 1 \times 10^{-3} \) for 50{,}000 epochs. L2 regularization with a coefficient of \( 1 \times 10^{-5} \) was applied to reduce overfitting. To balance the scales between \( \mathcal{L}_{\text{data}} \) and \( \mathcal{L}_{\text{ODE}} \), we set the weight \( w_{\text{ODE}} = 0.1 \). All hyperparameters were manually tuned to optimize predictive accuracy. Training and testing were conducted on an NVIDIA GeForce RTX 3090 GPU.

We trained the PINNs model using historical data from July 2020 to April 2022, employing a rolling window approach, as illustrated in Fig.~\ref{figure_pinn_schematic}B, to enable continuous predictions for the subsequent 1 to 4 weeks. Initially, PINNs were trained on data from weeks 1 to 17, with predictions made for weeks 18 to 21. The training window was then shifted by one week, retraining the model on data from weeks 1 to 18 and forecasting weeks 19 to 22. This process was repeated until PINNs were trained using data from weeks 1 to 89 and predictions were made for weeks 90 to 93. As a result, each week from 21 to 90 was treated as the subsequent 1-, 2-, 3-, and 4-week forecast in the rolling window procedure.

\subsection{Evaluation metrics}

We used two metrics to assess predictive performance: the mean absolute scaled error (MASE)~\cite{hyndman2006another} for point forecasting and the weighted interval score (WIS) for quantile forecasting. MASE was calculated by dividing the mean absolute error (MAE) of the model's predictions by the MAE of a naive model of the same forecast horizon.

\begin{equation}
    MASE = \frac{{MAE_{PINNs}}}{MAE_{naive}}
\end{equation}

MAE was calculated as the mean absolute difference between model predictions $\hat{y_i}$ and empirical observations $y_i$:

\begin{equation}
    MAE = \frac{1}{N}{\sum_{i=1}^{N}|\hat{y_i}-y_i|}
\end{equation}
where $N$ is the number of data.

The naive model uses data from the previous weeks as predictions for the following weeks. Specifically, when the forecast horizon is 1 week, the naive model uses data from the past week to predict the following week. Similarly, for a 2-week forecast horizon, the naive model uses data from the past 2 weeks to predict the subsequent 2 weeks. A MASE value smaller than 1.0 indicates that our PINNs model outperforms the naive model in terms of predictive accuracy. Moreover, a lower MASE value corresponds to more accurate predictions.

In infectious disease forecasting, quantile predictions, namely probabilistic predictions, are typically required to account for error propagation. We assume that the prediction distribution follows a Gaussian distribution, with mean \( \mu_t \) equal to the point forecast at time \( t \), and standard deviation \( \sigma \), which represents the standard deviation of all prediction errors and is constant across time. The prediction quantiles at time \( t \) are then given by \( \mu_t + \sigma \sqrt{2} \, \text{erf}^{-1}(2p - 1) \), where \( \text{erf}^{-1} \) denotes the inverse error function and \( p \) represents the desired quantile.

To evaluate probabilistic predictions using the WIS, we employed the `interval-scoring' Python package (\url{https://github.com/adrian-lison/interval-scoring}), based on~\cite{bracher2021evaluating}. We calculated the WIS with quantile bins as required by the COVID-19 Forecast Hub, specifically \( \alpha_1 = 0.02 \), \( \alpha_2 = 0.05 \), \( \alpha_3 = 0.1 \), \ldots, \( \alpha_{11} = 0.9 \), and quantiles including 0.010, 0.025, 0.050, 0.100, 0.150, 0.200, 0.250, 0.300, 0.350, 0.400, 0.450, 0.500, 0.550, 0.600, 0.650, 0.700, 0.750, 0.800, 0.850, 0.900, 0.950, 0.975, and 0.990.

\subsection{Ablation study}

In contrast to traditional neural networks, we incorporated epidemiological knowledge into our model by adding the ODE loss to the overall loss function. To assess the impact of embedding domain-specific epidemiological knowledge into the neural network, we also trained a traditional neural network (denoted as NN) and compared its performance with our PINNs model. The NN shares the same architecture as the neural network component in the upper left part of Fig.~\ref{figure_pinn_schematic}A, but it is purely data-driven and does not integrate the compartmental model.

In contrast to traditional mathematical methods, we employed a neural network and integrated it with the compartmental model. To evaluate the advantages of PINNs over the compartmental model, we compared the performance of our PINNs with that of the Gaussian Infection State Space with Time Dependence (GISST) model~\cite{drake2023data}. GISST is a conventional mathematical model that solves the same compartmental model.

It is important to note that early historical data may not accurately capture the dynamics of current virus variants due to mutations, and may even negatively impact future predictions. To assess this, we investigated the effect of the training data window size on prediction accuracy. While PINNs were originally trained using all available historical data, we also trained models using only recent data and compared their predictive performance. The results are presented in Appendix~\ref{appendix_1}.

Additionally, we conducted an analysis of the impact of L2 regularization on model performance and evaluated its necessity in our forecasting tasks. This analysis helps to assess whether regularization significantly contributes to improving generalization and preventing overfitting. The results of these experiments, including performance comparisons with and without L2 regularization, are presented in Appendix~\ref{appendix_2}.

\section{Results}

\subsection{Point forecasts on the number of cases, deaths, and hospitalizations for the following 1 - 4 weeks}

\begin{figure}[ht]
  \centering
\hspace{0.1in}\includegraphics[width=1.0\textwidth]{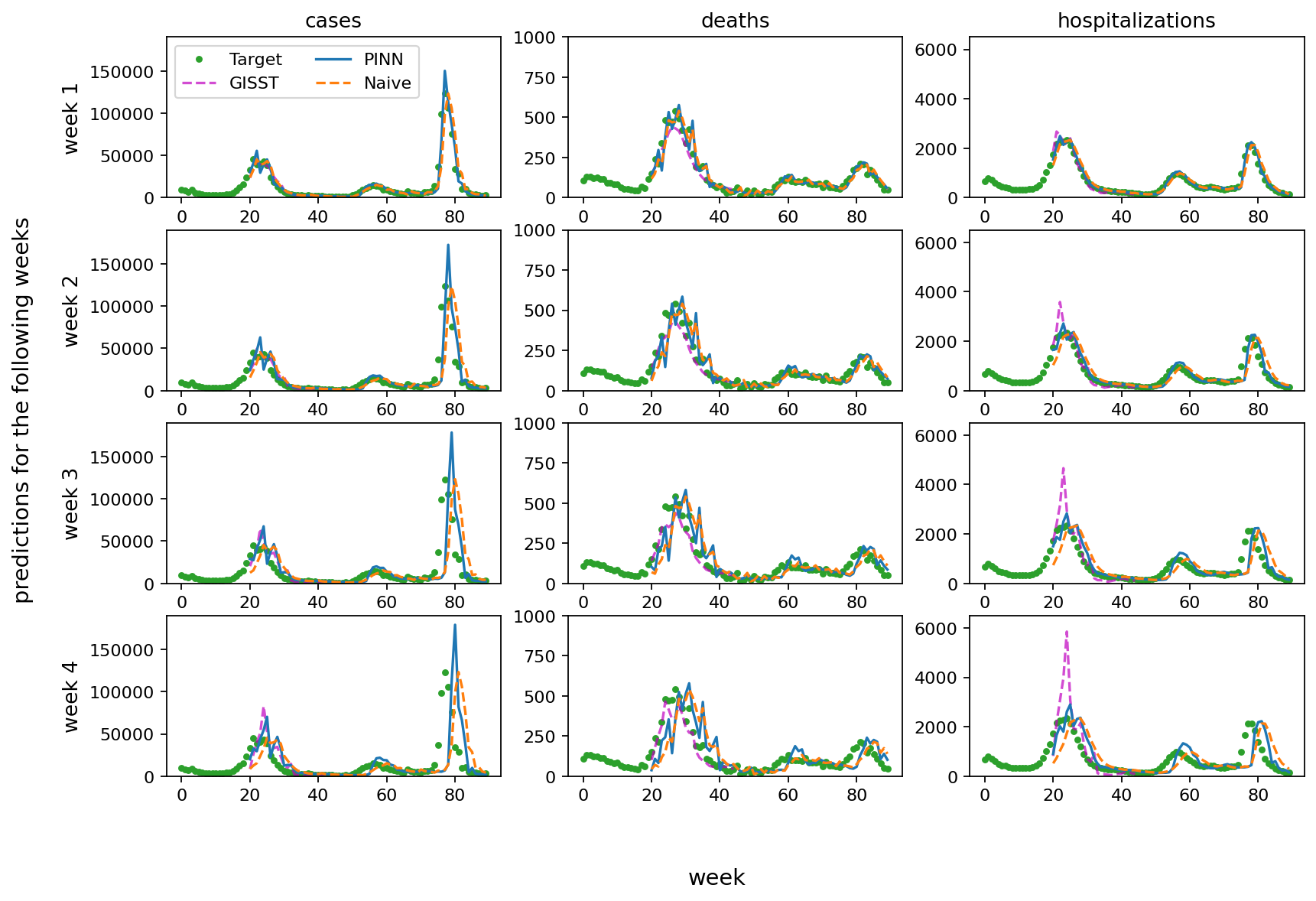}
  \caption{PINNs' point predictions on the number of cases, deaths, and hospitalizations for the following 1 - 4 weeks. GISST represents a mathematical model named Gaussian infection state space with time dependence~\cite{o2022semi}. The naive model uses data from the previous weeks as predictions for the following weeks. }
  \label{figure_result_pinn_1to4weeks}
\end{figure}

At each time point $t$, we trained PINNs using historical data from July 2020 up to $t$, and generated forecasts for the subsequent 1 to 4 weeks. The predictions, shown in Fig.~\ref{figure_result_pinn_1to4weeks}, demonstrate that the outputs from PINNs are consistently aligned with the observed target values and are more accurate than those from the naive model across all forecasting horizons. Additionally, PINNs outperform most of the corresponding forecasts produced by GISST. The results also indicate that one-week-ahead predictions are the most reliable, with accuracy gradually declining for forecasts on the subsequent 2-, 3-, and 4-week horizons. As the forecasting horizon increases, the predicted values tend to diverge more from the ground truth, accompanied by greater fluctuations in the predicted curves.

Quantitative comparisons are reported in Tab.~\ref{tab:table_mase_wis_comparison}. In the column labeled `Method', the name before the dash indicates the forecasting approach, while the number denotes the forecast horizon. Scores highlighted in red represent the best-performing model (the lowest score) for a given forecast horizon within each metric column. For example, `PINNs - 1 week' refers to PINNs predictions for the subsequent one week, while `PINNs - 2 week' refers to forecasts for the second week ahead. We observe that all of PINNs' MASE scores are no more than 1.0, indicating that PINNs consistently outperform the naive model across all forecast horizons. Among the three state variables, the MASE scores for deaths are generally higher than those for cases and hospitalizations, likely due to greater variability and fluctuations in the original death data.

\begin{table}[ht]
    \small
    \centering
        \begin{tabular}{|c|c|c|c|c|c|c|c|c|c|}
        \hline
        \multirow{2}{*}{Method} & \multicolumn{3}{c|}{MASE} & \multicolumn{3}{c|}{WIS} & \multicolumn{3}{c|}{scaled WIS}\\
        \cline{2-10} & Cases & Deaths & Hosp & Cases & Deaths & Hosp & Cases & Deaths & Hosp\\
        \hline
        PINN - 1 week & \textcolor{red}{0.75} & 1.00 &  \textcolor{red}{0.58} & 16247 & \textcolor{red}{103} & \textcolor{red}{250} & 0.70 & \textcolor{red}{1.04} & \textcolor{red}{0.57}\\
        \hline
        PINN - 2 week & 0.77 & 0.92 & \textcolor{red}{0.62} & 32954 & 146 & \textcolor{red}{556} & 0.79 & 1.00 & \textcolor{red}{0.66}\\
        \hline
        PINN - 3 week & 0.81 & 0.86 & \textcolor{red}{0.68} & 49272 & \textcolor{red}{176} & \textcolor{red}{882} & 0.87 & \textcolor{red}{0.91} & \textcolor{red}{0.72}\\
        \hline
        PINN - 4 week & 0.85 & 0.89 & \textcolor{red}{0.74} & 63311 & 217 & \textcolor{red}{1232} & 0.94 & 0.89 & \textcolor{red}{0.79}\\
        \hline\hline
        NN - 1 week & 0.91 & 1.32 & 0.65 & 25152 & 158 & 298 & 1.09 & 1.60 & 0.68\\
        \hline
        NN - 2 week & 1.27 & 1.76 & 0.79 & 52761 & 298 & 729 & 1.27 & 2.04 & 0.86\\
        \hline
        NN - 3 week & 1.56 & 1.86 & 0.89 & 88920 & 442 & 1281 & 1.57 & 2.29 & 1.05\\
        \hline
        NN - 4 week & 1.65 & 1.84 & 0.97 & 118515 & 558 & 1885 & 1.76 & 2.28 & 1.22\\
        \hline\hline
        GISST - 1 week & 0.85 & \textcolor{red}{0.88} & 0.91 & \textcolor{red}{15346} & 163 & 401 & \textcolor{red}{0.66} & 1.65 & 0.92 \\
        \hline
        GISST - 2 week & \textcolor{red}{0.75} & \textcolor{red}{0.64} & 0.78 & \textcolor{red}{26233} & 185 & 798 & \textcolor{red}{0.63} & 1.27 & 0.94 \\
        \hline
        GISST - 3 week & \textcolor{red}{0.71} & \textcolor{red}{0.45} & 0.77 & \textcolor{red}{37041} & 187 & 1300 & \textcolor{red}{0.65} & 0.97 & 1.06 \\
        \hline
        GISST - 4 week & \textcolor{red}{0.76} & \textcolor{red}{0.34} & 0.89 & 58148 & \textcolor{red}{181} & 2159 & 0.86 & \textcolor{red}{0.74} & 1.39 \\
        \hline\hline
        RNN - 1 week & 0.91 & 1.28 & 0.70 & 17653 & 126 & 278 & 0.76 & 1.27 & 0.64 \\
        \hline
        RNN - 2 week & 0.92 & 1.17 & 0.82 & 33544 & 168 & 642 & 0.81 & 1.15 & 0.76 \\
        \hline
        RNN - 3 week & 0.94 & 1.09 & 0.94 & 46641 & 205 & 1105 & 0.82 & 1.06 & 0.90 \\
        \hline
        RNN - 4 week & 0.92 & 0.99 & 1.03 & \textcolor{red}{53964} & 242 & 1554 & \textcolor{red}{0.80} & 0.99 & 1.00 \\
        \hline\hline
        LSTM - 1 week & 1.00 & 1.29 & 0.79 & 19622 & 124 & 325 & 0.85 & 1.25 & 0.74 \\
        \hline
        LSTM - 2 week & 1.00 & 1.39 & 0.93 & 35249 & 194 & 786 & 0.85 & 1.33 & 0.93 \\
        \hline
        LSTM - 3 week & 0.98 & 1.16 & 1.03 & 49661 & 226 & 1249 & 0.87 & 1.17 & 1.02 \\
        \hline
        LSTM - 4 week & 0.94 & 1.09 & 1.04 & 56474 & 268 & 1624 & 0.84 & 1.09 & 1.05 \\
        \hline\hline
        GRU - 1 week & 0.89 & 1.22 & 0.77 & 16952 & 115 & 324 & 0.73 & 1.16 & 0.74 \\
        \hline
        GRU - 2 week & 0.90 & 1.24 & 0.88 & 33573 & 178 & 749 & 0.81 & 1.22 & 0.88 \\
        \hline
        GRU - 3 week & 0.93 & 1.14 & 0.98 & 47267 & 225 & 1198 & 0.83 & 1.17 & 0.98 \\
        \hline
        GRU - 4 week & 0.90 & 1.03 & 1.01 & 55414 & 262 & 1608 & 0.82 & 1.07 & 1.04 \\
        \hline\hline
        Transformer - 1 week & 0.96 & 1.14 & 0.81 & 22155 & 111 & 356 & 0.96 & 1.12 & 0.81 \\
        \hline
        Transformer - 2 week & 0.99 & 0.98 & 0.86 & 40494 & \textcolor{red}{145} & 730 & 0.97 & \textcolor{red}{0.99} & 0.86 \\
        \hline
        Transformer - 3 week & 1.05 & 0.97 & 0.94 & 56804 & 193 & 1179 & 1.00 & 1.00 & 0.96 \\
        \hline
        Transformer - 4 week & 1.08 & 0.91 & 1.04 & 69297 & 235 & 1629 & 1.03 & 0.96 & 1.05 \\
        \hline
    \end{tabular}
    \caption{MASE and WIS comparison of different methods for point forecasting and quantile forecasting, respectively. Column Hosp denotes hospitalizations. Scaled WIS is computed by dividing the WIS by the WIS of naive models of the same forecast horizon. Scores in red denote the lowest score of the same forecast horizon in that column. }
    \label{tab:table_mase_wis_comparison}
\end{table}

\subsection{Quantile forecasts on the number of cases, deaths, and hospitalizations for the following 1 - 4 weeks}

\begin{figure}[ht]
  \centering
\hspace{0.1in}\includegraphics[width=0.95\textwidth]{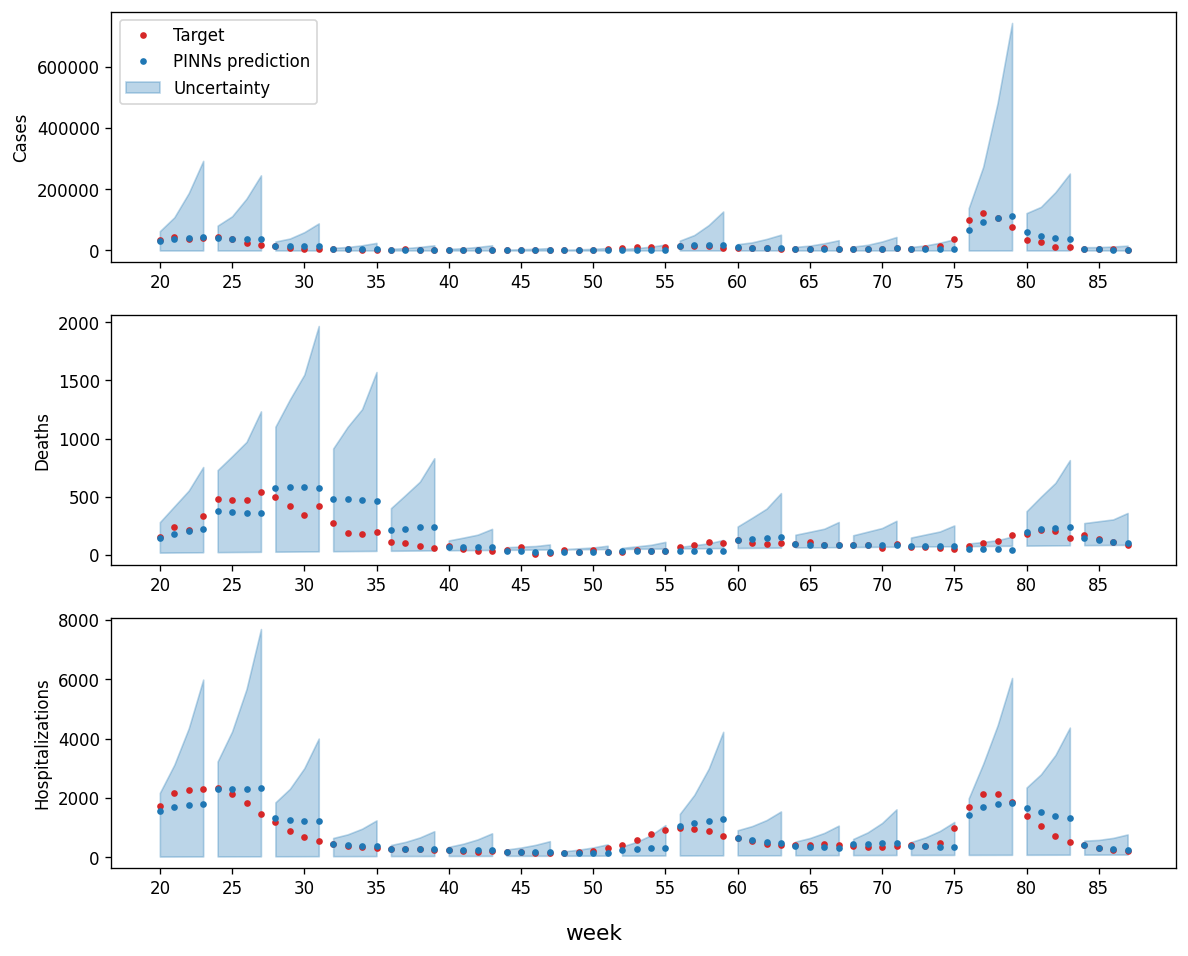}
  \caption{PINNs' quantile predictions on the number of cases, deaths, and hospitalizations for the following 1 - 4 weeks. The red dots represent the ground truth. The blue points indicate PINNs' predictions and the light blue region represents the associated uncertainty in the forecasts. Each polygon spans a continuous four-week prediction interval, encompassing the 1-, 2-, 3-, and 4-week forecasts made from a time point one week prior to the start of the polygon }
  \label{figure_result_pinn_1to4weeks_quantile}
\end{figure}

Similarly, at each time point $t$, we trained PINNs using historical data from July 2020 up to $t$ and generated predictions for the subsequent 1 to 4 weeks. In this case, we visualized the continuous forecasts along with the associated uncertainty in Fig.~\ref{figure_result_pinn_1to4weeks_quantile}. For each group of 1- to 4-week predictions, the points (representing the median predictions at quantile = 0.5) progressively deviate from the ground truth as the forecast horizon extends. Correspondingly, the light blue region, which represents predictive uncertainty, expands with longer horizons. This pattern highlights a common trend: as the forecast horizon increases, prediction accuracy decreases while uncertainty grows.

Quantitative comparisons of WIS are presented in Tab.~\ref{tab:table_mase_wis_comparison}. The scaled WIS was computed by dividing the WIS of each model by the WIS of the naive model for the corresponding forecast horizon. We observe that PINNs achieve scaled WIS scores that are generally below 1.0, indicating superior performance over the naive baseline across most forecasting horizons. The only exception occurs in the 1-week-ahead quantile predictions for death, where the PINNs' WIS score slightly exceeds 1.0. However, this deviation is minor and still significantly lower than the scores of other models.

\subsection{Comparison with sequence deep learning models and traditional mathematical model}

 We first compared our PINNs predictions with results obtained from the traditional neural network model (NN) without incorporating the compartmental model as well as with the results of a traditional mathematical model, named GISST~\cite{o2022semi}. As shown in Tab.~\ref{tab:table_mase_wis_comparison}, both MASE and WIS scores for PINNs are lower than the corresponding scores of NN for each forecast horizon, which demonstrates the absolute advantage of PINNs over traditional neural networks for both point forecasting and quantile forecasting. This is likely due to the regularizing effect of the compartmental model.
 
 Additionally, PINNs typically outperformed GISST when forecasting hospitalizations, GISST outperformed PINNs for cases, and forecasting performance for deaths depended on whether the evaluation was for point forecasts (GISST superior) or quantile forecasts (PINNs superior). These results demonstrate the important role of epidemiological domain knowledge as a constraint to avoid unreasonable predictions that may arise in purely data-driven modeling.

To further evaluate the performance of PINNs, we compared them against several state-of-the-art neural network architectures commonly used for sequential data processing. Specifically, we implemented RNNs~\cite{grossberg2013recurrent},  LSTMs~\cite{hochreiter1997long}, GRUs~\cite{dey2017gate}, and Transformer models~\cite{ghojogh2020attention} to address the same forecasting task. The RNN, LSTM, and GRU models share a similar architecture configuration. Each model consists of an input layer with three neurons, two hidden layers with 50 neurons each, and an output layer with three neurons. The models take weekly data on cases, deaths, and hospitalizations as input features and generate corresponding predictions for each of the three variables. A window size of seven weeks was used for input. Training was conducted using the Adam optimizer with a learning rate of $1 \times 10^{-3}$ over 10{,}000 epochs.

The Transformer model is composed of an input layer with three neurons, followed by two encoder layers, each with a hidden dimensionality of 64 and four attention heads, and an output layer with three neurons. Similar to the other models, it takes weekly data on cases, deaths, and hospitalizations as input features and outputs corresponding predictions for each variable. A window size of seven weeks was used to form the input sequence. The model was trained using the Adam optimizer with a learning rate of $1 \times 10^{-3}$ for 10,000 epochs.

\begin{figure}[ht]
  \centering
\hspace{0.1in}\includegraphics[width=1.0\textwidth]{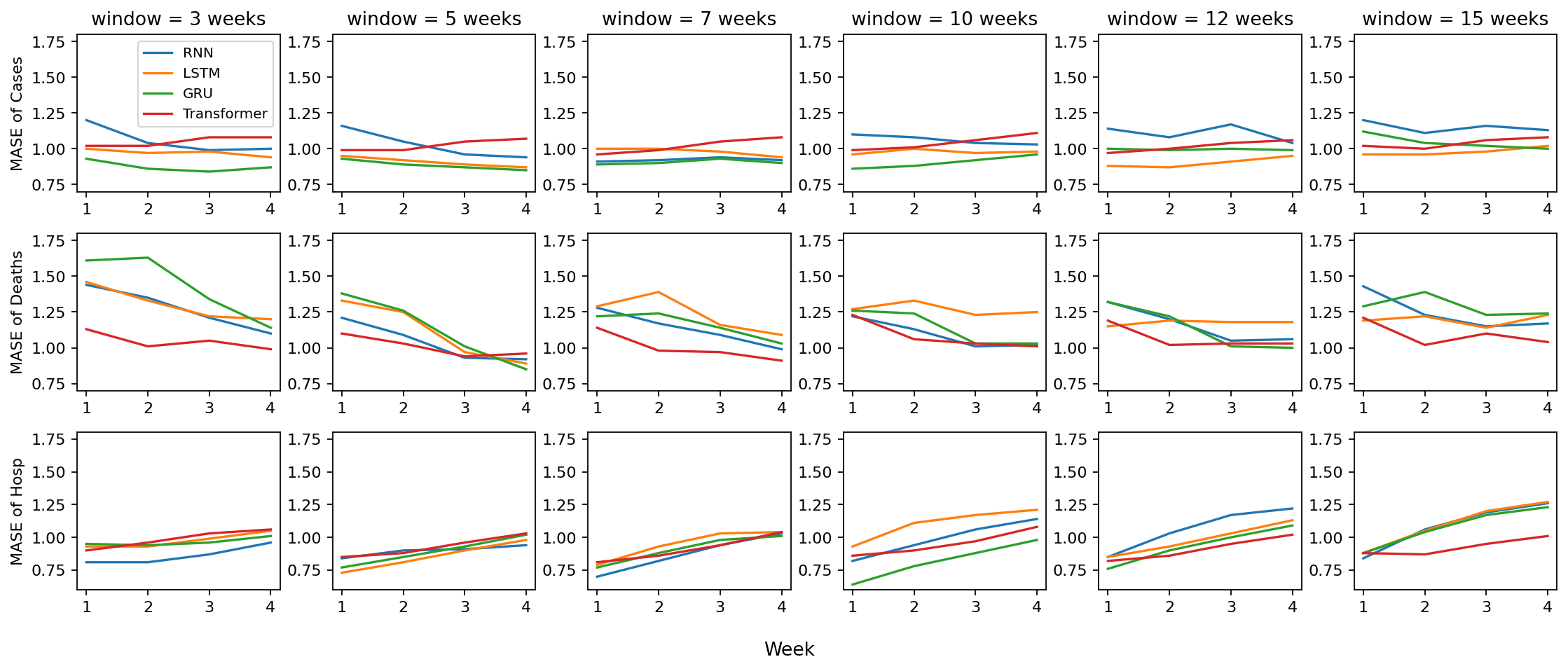}
  \caption{A MASE comparison of Recurrent Neural Networks (RNNs), Long Short-Term Memory (LSTM) networks, Gated Recurrent Units (GRUs), and Transformer models, with training window sizes ranging from 3 weeks to 15 weeks.} 
  \label{figure_result_other_models_window_size}
\end{figure}

The window size is a crucial hyperparameter for these models. Thus, we evaluated the models using window sizes of 3, 5, 7, 10, 12, and 15 weeks, and the results are presented in Fig.~\ref{figure_result_other_models_window_size}. The findings align with our expectations: a window that is too short does not adequately capture the temporal dependencies of the data, whereas an excessively long window introduces redundancy. A window size of seven weeks strikes an optimal balance and is well-suited for this forecasting task.

\begin{figure}[ht]
  \centering
\hspace{0.1in}\includegraphics[width=1.0\textwidth]{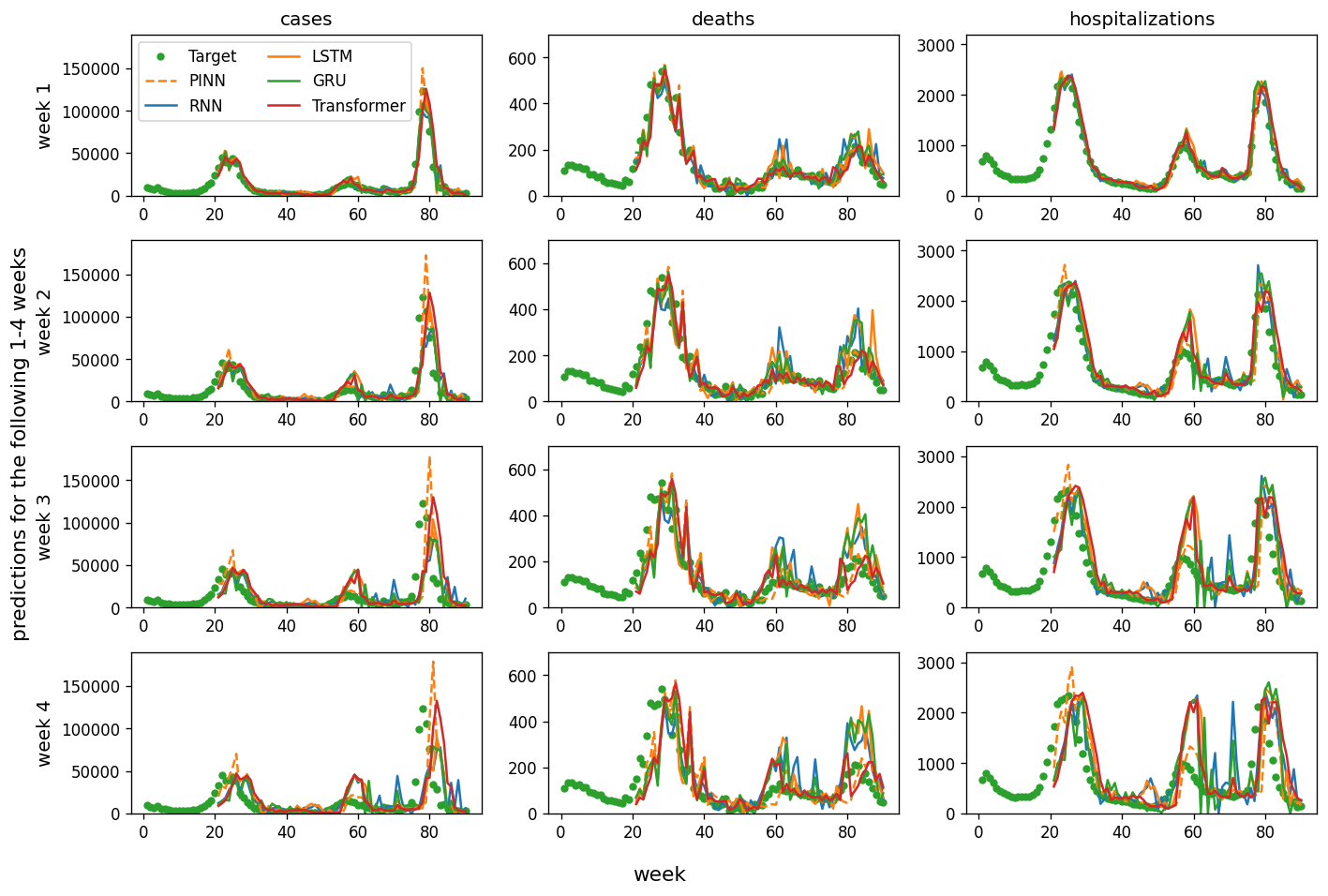}
  \caption{MASE comparison with Recurrent Neural Networks (RNNs), Long Short-Term Memory (LSTM) networks, Gated Recurrent Units (GRUs), and Transformers on the number of cases, deaths, and hospitalizations for the following 1 - 4 weeks. } 
  \label{figure_result_other_models_predictions}
\end{figure}

The predictions for one-, two-, three- and four-week forecast are shown in Fig.~\ref{figure_result_other_models_predictions}. In comparison to PINNs, the forecasts generated by RNNs, LSTMs, GRUs, and Transformers exhibit noticeable fluctuations, indicating a higher degree of overfitting. This contrast underscores the advantage of incorporating physical information into PINNs, which contributes to more stable and reliable predictions. A quantitative comparison is provided in Tab.~\ref{tab:table_mase_wis_comparison}, where PINNs consistently outperform all other prominent neural network models commonly used for time series forecasting, both in terms of MASE for point forecasting and WIS for quantile forecasting. Despite architectural advances in RNNs, LSTMs, GRUs, and Transformers, none of these models demonstrates substantial performance gains on this task. These findings emphasize the importance of integrating physics-based domain knowledge into neural network models for improved generalization and predictive accuracy.

\section{Discussion and Summary}

Epidemic forecasting can traditionally be framed as an inverse-forward problem, wherein model parameters must first be inferred before generating future predictions. In many scenarios, the mechanisms underlying disease transmission are partially understood and represented using compartmental models in the form of ODEs. However, several model parameters remain unknown and must be estimated by fitting the model to observational data. Once the parameters are identified, these compartmental models can be employed to forecast future trends in disease spread and evaluate the impact of public health interventions or the emergence of new variants.

Despite their utility, traditional mathematical models suffer from inherent limitations due to model misspecification. For example, they often assume fixed transition rates, which makes it challenging to adapt to evolving conditions, such as changes in transmission dynamics caused by new variants. Moreover, forecasting is not feasible unless all relevant parameters are accurately estimated. In contrast, data-driven neural network models have emerged as powerful tools for addressing complex social and scientific problems in recent years. However, conventional neural networks typically require large volumes of training data and labeled examples to perform effectively, and they often disregard prior domain knowledge. As a result, machine learning models may produce unrealistic or unreliable forecasts, particularly in extrapolative scenarios, due to their lack of grounding in the underlying physical or epidemiological processes.

In this work, we employed PINNs to integrate epidemiological knowledge, as described by compartmental models, with observed data to enhance the accuracy of infectious disease forecasts. Our results demonstrate that the proposed approach offers several key advantages that contribute to its superior performance. First, the PINN framework leverages neural networks to model complex and non-linear patterns inherent in infectious disease dynamics, enabling the system to capture intricate relationships that traditional models may miss. Second, PINNs incorporate domain-specific knowledge by embedding the compartmental models into the loss function. This integration ensures that the learned representations conform to the underlying principles and constraints of disease transmission. Moreover, the inclusion of mechanistic models in the training process helps mitigate overfitting, a common issue when relying solely on data-driven machine learning approaches. By constraining the learning process with known epidemiological laws, PINNs achieve a balanced trade-off between flexibility and generalizability.

Additionally, PINNs can be employed to estimate key epidemiological parameters—such as transmission and recovery rates -- directly from data, thereby enhancing the model's adaptability to evolving disease dynamics. In particular, PINNs treat these parameters as outputs of the neural network, allowing them to vary over time. This enables the model to capture temporal changes in transmission behavior, such as those driven by viral mutations or shifts in public health interventions. Furthermore, PINNs are capable of continuously incorporating newly acquired data, allowing for real-time model updates and progressively improved forecasts as more information becomes available during an ongoing outbreak. This adaptability makes PINNs particularly well-suited for real-world epidemic surveillance and response. In addition, PINNs can integrate heterogeneous data sources -- including social, environmental, and demographic factors -- offering a more comprehensive and holistic framework for infectious disease forecasting.

Recently, several PINNs-based models have been developed for infectious disease forecasting, including applications to COVID-19~\cite{berkhahn2022physics, rodriguez2023einns, barmparis2022physics} and measles~\cite{madden2024neural}. Compared to these existing approaches, our proposed model incorporates a more complex compartmental framework with nine state variables embedded into the neural network architecture. This significantly increases the challenge of ensuring consistency with the underlying physics described by the ODE system, thereby demonstrating the flexibility and robustness of our model. Notably, our model achieved better performance metrics -- Normalized Root Mean Squared Error (NR1 and NR2) and Normal Deviation (ND), as defined in~\cite{rodriguez2023einns} -- compared to the Epidemiologically-Informed Neural Networks (EINNs)~\cite{rodriguez2023einns} in forecasting COVID-19 deaths, which is the only predicted state variable in EINNs. Specifically, our model achieved 0.53 (EINNs: 0.54) on NR1, 0.15 (EINNs: 0.24) on NR2, and 0.33 (EINNs: 0.38) on ND, while simultaneously predicting two additional state variables. Furthermore, while prior models primarily offer point forecasts, we extend our approach to quantile forecasts, which allows us to quantify uncertainty in the predictions.

Despite the strengths of the vanilla PINN framework in handling noisy and limited data, it lacks a built-in mechanism for quantifying the uncertainty of its predictions in the presence of input noise. A well-established method for regression with uncertainty estimation is Gaussian Process Regression (GPR), which, rooted in the Bayesian framework, provides a principled way to quantify uncertainty in the solution of differential equations. However, GPR is typically limited to small, linear problems due to its high computational complexity, which scales poorly with large datasets. A promising direction to address this limitation involves the use of Bayesian PINNs, as proposed by~\cite{yang2021b}. These models combine the physics-informed learning capability of PINNs with the probabilistic nature of Bayesian inference, allowing for uncertainty-aware solutions even in ill-posed settings, such as those with incomplete or under-specified boundary conditions. Future research should investigate the applicability and effectiveness of Bayesian PINNs in the context of infectious disease forecasting, particularly in scenarios involving high uncertainty, evolving dynamics, or limited data availability.

In summary, the application of PINNs in infectious disease forecasting offers a synergistic approach that combines the power of data-driven deep learning with the interpretability and constraints provided by underlying epidemiological principles, leading to more accurate, interpretable, and adaptable models for predicting the spread of infectious diseases.

\section*{Data Availability}
All relevant code and data are available from \url{https://github.com/dpdclub/PINNs-for-Epidemiology-}.

\section*{Funding}
This work was supported by National Institute of Health grants  R21HL168507 and NSF SCH Award Number:2406212. High-performance computing resources were provided by The Georgia Advanced Computing Resource Center (GACRC) at the University of Georgia. 

\section*{Acknowledgments}
H.L. would like to thank Shan-Ho Tsai from the Georgia Advanced Computing Resource Center (GACRC) for providing technical support.

\section{Appendices}

\subsection{Impact of training data length}\label{appendix_1}

\begin{figure}[ht]
  \centering
\hspace{0.1in}\includegraphics[width=1.0\textwidth]{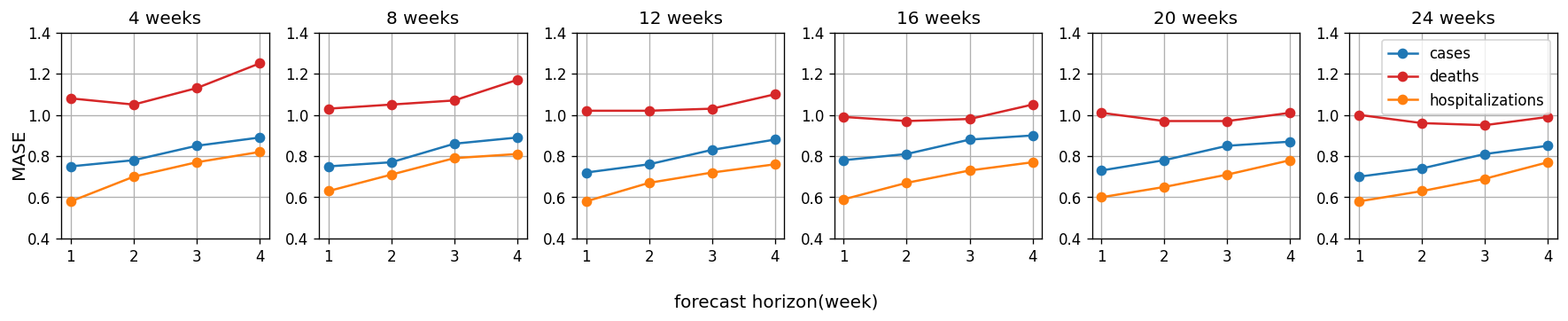}
  \caption{MASE for the predictions of PINNs models on cases, deaths, and hospitalizations when trained with varied lengths of historical data ranging from 4 weeks to 24 weeks.} 
  \label{figure_result_training_data_length}
\end{figure}

We trained PINNs using datasets comprising the most recent 4, 8, 12, 16, 20, and 24 weeks of observations, and evaluated their forecasting performance over the period spanning weeks 48 to 110. The results, shown in Fig.~\ref{figure_result_training_data_length}, illustrate the MASE scores for predictions of cases, deaths, and hospitalizations across 1-, 2-, 3-, and 4-week horizons using different training data lengths.

For all three targets, we observe a consistent trend: as the length of training data increases, the MASE scores generally decrease and then stabilize. Models trained on only 4 or 8 weeks of data tend to underperform, likely due to insufficient information to accurately capture the underlying disease dynamics. In contrast, models trained on 12 or 16 weeks of data achieve predictive accuracy comparable to those trained on the full historical dataset. Notably, further increasing the training window to 20 or 24 weeks yields only marginal improvements in accuracy. This suggests that data beyond 16 weeks—spanning nearly half a year—may be less relevant to current epidemiological conditions and contribute less to near-term forecasting performance.

\subsection{Impact of L2 regularization}\label{appendix_2}

We further demonstrate that the absence of L2 regularization leads to a substantial decline in model accuracy. Specifically, removing L2 regularization results in average performance drops of 22.48\% for cases, 9.16\% for deaths, and 42.54\% for hospitalizations over the 1- to 4-week forecasting horizons. These results highlight the importance of incorporating L2 regularization in the PINNs framework, as it contributes to improved generalization and prediction stability across multiple epidemiological targets.

\bibliographystyle{unsrt}
\bibliography{ref}

\end{document}